\documentclass[10pt,twocolumn,letterpaper]{article}

\usepackage{iccv}
\usepackage{times}
\usepackage{epsfig}
\usepackage{graphicx}
\usepackage{amsmath}
\usepackage{amssymb}

\usepackage{axessibility}  
\usepackage{wrapfig}
\usepackage{float}
\usepackage[super]{nth}
\usepackage{bm}
\usepackage{makecell}
\usepackage{boldline}
\usepackage{multirow}
\usepackage{amsmath}
\usepackage{threeparttable}
\usepackage{array}
\usepackage{algorithm,algpseudocode}
\usepackage{color}
\usepackage{enumerate}
\usepackage{subfigure}
\usepackage[super]{nth}
\usepackage{paralist}
\usepackage{booktabs}

\usepackage[pagebackref=true,breaklinks=true,letterpaper=true,colorlinks,bookmarks=false]{hyperref}

\iccvfinalcopy 

\def\iccvPaperID{4000} 

\ificcvfinal\fi

\begin{document}

\title{Meta-Aggregator: Learning to Aggregate for 1-bit Graph Neural Networks}

\author{
Yongcheng Jing$^1$,
Yiding Yang$^2$,
Xinchao Wang$^{3}$,
Mingli Song$^4$,
Dacheng Tao$^{5,1}$
 \\ 
$^1$The University of Sydney, Australia, 
$^2$Stevens Institute of Technology, \\
$^3$National University of Singapore, 
$^4$Zhejiang University,
$^5$JD Explore Academy, China\\
{\tt\small
yjin9495@sydney.edu.au, 
yyang99@stevens.edu,}\\
{\tt\small
xinchao@nus.edu.sg,
brooksong@zju.edu.cn,
dacheng.tao@gmail.com
}
}

\maketitle
\ificcvfinal\thispagestyle{empty}\fi

\begin{abstract}
In this paper, we study a novel meta aggregation scheme 
towards binarizing graph neural networks (GNNs).
We begin by developing a vanilla 1-bit GNN framework that
binarizes both the GNN parameters and the graph features.
Despite the lightweight architecture, 
we observed that this vanilla framework suffered from insufficient discriminative power 
in distinguishing graph topologies, 
leading to a dramatic drop in performance.
This discovery motivates us to devise meta aggregators to 
improve the expressive power of vanilla binarized GNNs, 
of which the aggregation schemes can be adaptively changed 
in a learnable manner based on the binarized features.
Towards this end, we propose two dedicated forms of meta neighborhood aggregators,  
an exclusive meta aggregator termed as {Greedy Gumbel Neighborhood Aggregator} (GNA),
and a diffused meta aggregator termed as 
{Adaptable Hybrid Neighborhood Aggregator} (ANA).
GNA learns to exclusively pick one single optimal aggregator from a pool of candidates,
while ANA learns a hybrid aggregation behavior to 
simultaneously retain the benefits of several individual aggregators.
Furthermore, the proposed meta aggregators may readily serve as
a generic plugin module into existing full-precision GNNs.
Experiments across various domains
demonstrate that the proposed method yields 
results superior to the state of the art. 

\end{abstract}

\thispagestyle{empty}
\pagestyle{empty}

\section{Introduction}

Graph neural networks (GNNs) have recently 
emerged as the dominant paradigm for learning and analyzing  
non-Euclidean data,
which contain rich node content information as well as topological relational information~\cite{dwivedi2020benchmarking,hu2020open,wu2020comprehensive}.
As such, a massive number of  GNN architectures have been developed \cite{kipf2017semi,velivckovic2018graph,xu2018powerful,yang2021spagan,zhou2018graph}.
The success of GNNs also triggers a great surge of interest in applying elaborated graph networks to various tasks across many domains, such as object detection \cite{hu2018relation,gu2018learning}, pose estimation \cite{yang2021learning}, point cloud processing \cite{landrieu2018large,wang2019dynamic,qi20173d},  and visual SLAM \cite{sarlin2020superglue}.
These GNN-based applications, in general, rely on cumbersome graph architectures to deliver gratifying results.
For example, SuperGlue, a GNN-based feature matching approach, requires 12M network parameters to achieve the state-of-the-art performance \cite{sarlin2020superglue}.

In practice, however, such applications typically require a compact and lightweight architecture for real-time interaction, especially in resource-constrained environments.
In the case of autonomous driving \cite{milz2018visual}, for example, it is critical to maintain fast and timely responses for GNN-based SLAM algorithms to handle complex traffic conditions, thereby leading to the urgent need of compressing cumbersome GNN models.
The work of \cite{yang2020distillating}, as the first attempt, 
leverages knowledge distillation to learn a compact student GNN with fewer parameters.
In spite of the improved efficiency, 
this approach still relies on the expensive floating-point operations, 
let alone a well-performed teacher model
pre-trained in the first place.

\begin{figure}[t]
    \centering
    \includegraphics[width=0.475\textwidth]{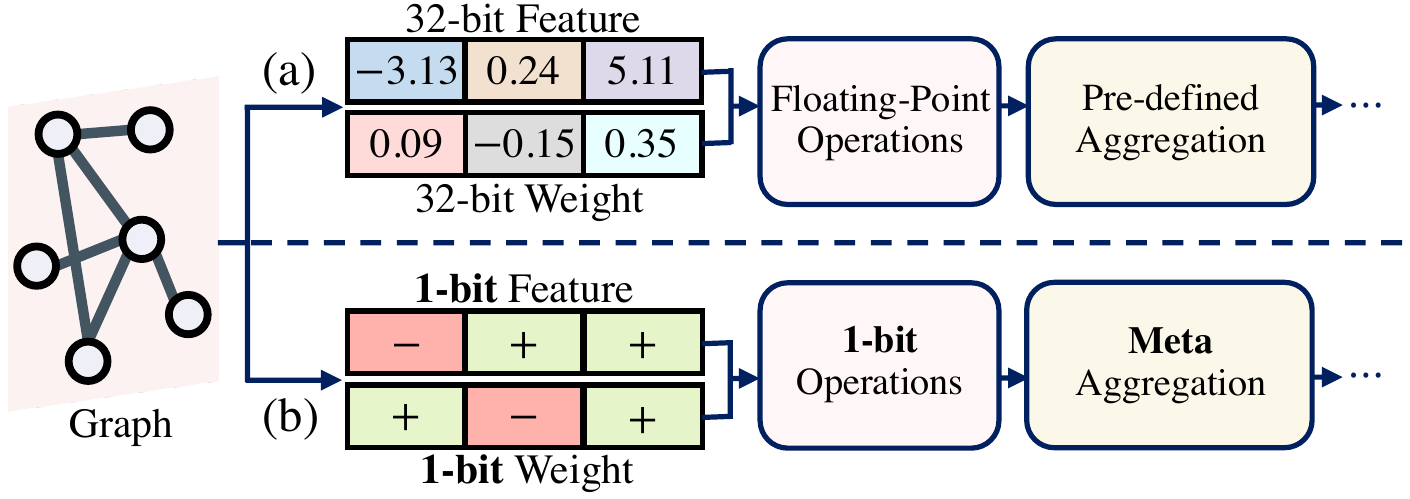}
    \caption{Illustrations of the computational workflow in (a) conventional full-precision GNNs and (b) the proposed 1-bit GNNs. In particular, we devise two meta aggregators for the proposed model, termed as \emph{Greedy Gumbel Aggregator}~(GNA) and \emph{Adaptable Hybrid Aggregator}~(ANA), that learn to perform adaptive aggregation in a graph-aware and layer-aware manner.}
    \label{fig:intro}
\end{figure}

\begin{figure}[t]
    \centering
    \includegraphics[width=0.475\textwidth]{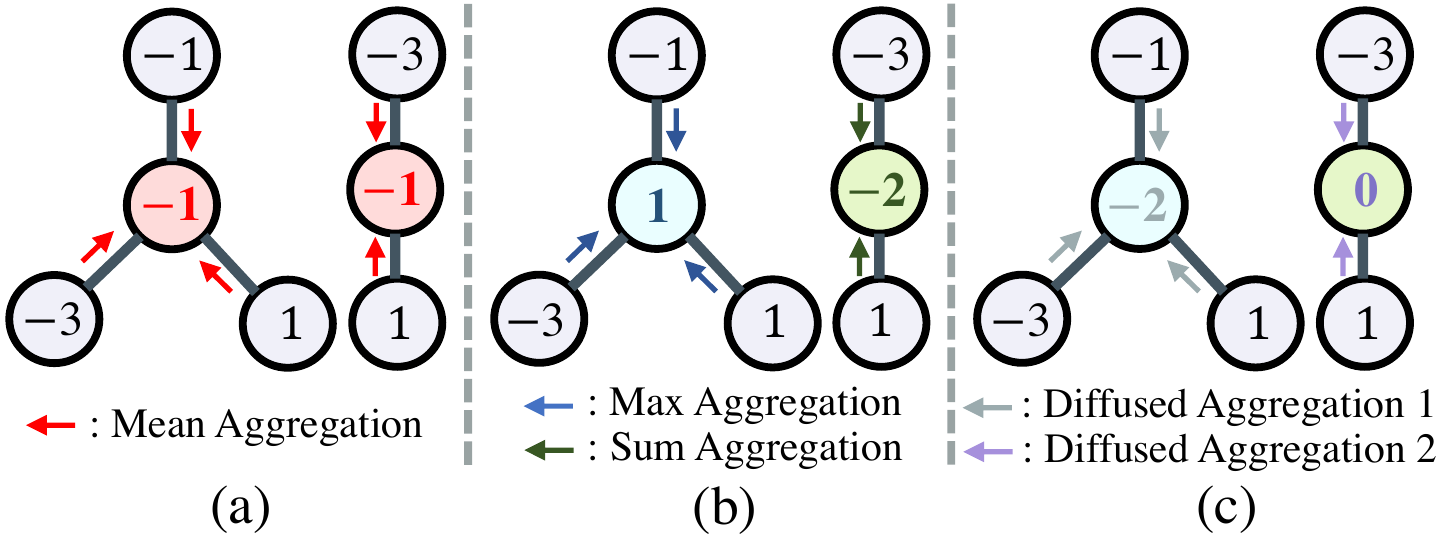}
    \caption{Example aggregation results of the two graphs with different topological structures for (a) the conventional pre-defined and fixed aggregator, (b) the proposed exclusive form of meta aggregators GNA, and (c) the proposed diffused form of meta aggregators ANA.}
    \label{fig:intro2}
    \vspace{-1em}
\end{figure}

In this paper, we strive to make one step further towards ultra lightweight GNNs.
Our goal is to train  a customized 1-bit GNN,
as shown in Fig.~\ref{fig:intro},
that allows for favorable memory efficiency and meanwhile enjoys competitive performance.
We start with developing a na\"ive GNN binarization framework, 
achieved through converting 32-bit features and parameters into 1-bit ones,
followed by leveraging straight-through estimator to optimize the binarized model.
The derived vanilla binarized GNN enjoys favorable memory efficiency;
however, its performance is not encouraging as expected. 
Through parsing its underlying process, we identified that 
the binarization  yields
limited expressive power, making the model
incapable to distinguish
different graph topologies.
An illustrating example is shown in  Fig.~\ref{fig:intro2}(a), 
where a \emph{mean} aggregator, which is commonly adopted by full-precision GNNs,
produce identical aggregation results
for two diversified graph topologies with binarized features, thereby leading to inferior performances.

Inspired by this discovery, we introduce to the 
proposed GNN binarization framework a learnable 
and adaptive neighborhood aggregator,
so as to alleviate the aforementioned dilemma 
and enhance the distinguishability
of 1-bit graphs.
Unlike existing GNNs that rely on a pre-defined and fixed aggregator, 
our elaborate meta neighborhood aggregators enables 
dynamically \emph{selecting} (Fig.~\ref{fig:intro2}(b)) or \emph{generating} (Fig.~\ref{fig:intro2}(c)) customized input- and layer-specific aggregation schemes.
As such, we explicitly account for the
customized characteristics of binarized graph features, and 
further strengthen the 
discriminative power for handling topological structures.

Towards this end, we propose two variants 
of meta aggregators:
an exclusive meta aggregator,
termed as \emph{Greedy Gumbel Neighborhood Aggregator} (GNA),
that adaptively \emph{selects} an optimal aggregator in a learnable manner,
as well as
a diffused meta aggregator,
termed as \emph{Adaptable Hybrid Neighborhood Aggregator} (ANA),
that  either \emph{approximates} a single
aggregator or dynamically \emph{generates} a hybrid aggregation behavior.
{Specifically, GNA incorporates the discrete decisions from the candidate aggregators, conditioned on the individual graph features, into the gradient descent process by leveraging \emph{Greedy Gumbel Sampling}.}
Inevitably, the performance of GNA is bottlenecked 
by the individual aggregators in the candidate pool.
Thus, we further devise ANA that enables
generating a hybrid aggregator dynamically
based on the input 1-bit graphs.
{ANA  simultaneously preserves the strengths of multiple individual aggregators, leading to favorable competence to handle the challenging 1-bit graph features.}
Moreover, the proposed GNA and ANA can  be readily extended as portable modules 
into the general full-precision GNN models to enhance the expressive capability.

In sum, our contribution is a novel GNN-customized binarization 
framework that  generates a 1-bit lightweight GNN model with competitive performance, making it competent for resource-constrained applications such as edge computing.
This is specifically achieved through an adaptive meta aggregation scheme to accommodate the challenging quantized graph features.
We evaluate the proposed customized framework on 
several large-scale benchmarks across
different domains and graph tasks.
Experimental results demonstrate that the proposed meta aggregators achieve results superior to the state-of-the-art, not only on the devised 1-bit binarized GNN models, but also on the general full-precision models.

\section{Related Work}


\noindent\textbf{Graph Neural Networks.} 
The concept of graph neural networks was proposed in \cite{scarselli2008graph}, which generalized existing neural networks to handle graph data represented in the non-Euclidean domain.
Over the past few years, graph neural networks have achieved unprecedented advances with various approaches being developed \cite{kipf2017semi,jing2021amalgamating,zhou2018graph,dwivedi2020benchmarking,yang2020factorizable,yan2018spatial,li2018adaptive,ma2019disentangled,wang2018graphgan,huang2018adaptive,li2018deeper,nt2019revisiting,liu2021overcoming}.
For example, graph attention network in \cite{velivckovic2018graph} introduces a novel attention mechanism for efficient graph processing.  
GraphSAGE \cite{hamilton2017inductive}, on the other hand, addresses the scalability issues on large-scale graphs by sampling and aggregating feature representations from local neighborhoods.

The success of GNNs has also boosted the applications of graph networks in a wide range of problem domains \cite{zhou2018graph}, including semantic segmentation \cite{dgcnn,landrieu2018large,qi20173d,qi2017pointnet++}, object detection \cite{hu2018relation,gu2018learning}, pose estimation \cite{yang2021learning}, interaction detection \cite{qi2018learning,jain2016structural}, and visual SLAM \cite{sarlin2020superglue}, \etc.
Specifically, Wang \etal \cite{dgcnn} propose a dynamic graph convolutional model for point cloud classification and semantic segmentation, which combines the advantages of the PointNet \cite{qi2017pointnet} and graph convolutional network \cite{kipf2017semi}.
Despite the encouraging performance, there is a lack of research on compressing cumbersome GNN models, which is critical for deployment in resource-constrained environments like on the mobile-terminal side.

We also notice two related works in \cite{li2020deepergcn,pellegrini2021learning} upon publication that focus on generalized aggregation functions.
However, our work is conceptually very different from \cite{li2020deepergcn,pellegrini2021learning}: 
in fact, our work is the first dedicated study on \emph{functionals} of GNNs, dealing with \emph{functions of functions}; in other words, we focus on learning {aggregators of aggregators}, where the inputs are themselves aggregators.
This has not yet been explored in prior works.

\noindent\textbf{Network Binarization.}
In the field of model compression \cite{yu2017compressing,shen2019amalgamating,shen2021progressive,chen2020addernet,shen2019customizing},
network binarization techniques aim to save memory occupancy and accelerate the network inference by binarizing network parameters and then utilizing bitwise operations \cite{hubara2016binarized,hubara2016binarizedarxiv,bulat2019xnor}.
In recent years, various CNN binarization methods have been proposed, which can be categorized into direct binarization \cite{courbariaux2015binaryconnect,hubara2016binarized,hubara2016binarizedarxiv,kim2016bitwise} and optimization-based binarization \cite{rastegari2016xnor,bulat2019xnor,martinez2020training}.
Specifically, direct binarization quantizes the weights and activations to 1 bit with a pre-defined binarization function.
In contrast, optimization-based binarization introduces scaling factors for the binarized parameters to improve the representation ability, but inevitably leading to inferior efficiency.

Driven by the success of the aforementioned binarization techniques in the CNN domain, in the paper, we propose a GNN-specific binarization method. 
Specifically, we primarily focus on GNN-based direct binarization, since our goal is to develop super lightweight GNN models. 
We also notice three concurrent works \cite{wang2020binarized,wang2020bi,bahri2020binary} that also aim to accelerate the forward process for GNN models. 
However, \cite{wang2020binarized,wang2020bi} directly apply CNN-based binarization techniques without considering the characteristics of GNNs, which in fact will serve as the baseline method in our experiments.
The other work in \cite{bahri2020binary} only focuses on improving the efficiency of dynamic graph convolutional model \cite{dgcnn}, by speeding up the dynamic construction of k-nearest-neighbor graphs in the Hamming space.   
Unlike \cite{wang2020binarized,wang2020bi,bahri2020binary}, we aim to devise a more general GNN-specific binarization framework that is applicable to most existing GNN models.

\section{Vanilla Binary GNN and Pre-analysis}
\label{sect:analysis}

In this section, we first develop a vanilla binary GNN framework by simply binarizing model parameters and activations.
We then show the limitations of this vanilla binary GNN by looking into the internal message aggregation process and accordingly develop two possible solutions to address these limitations.
Eventually, built upon the possible solutions, we introduce the idea of the proposed customized GNN binarization framework with the meta aggregators.

\vspace{-1em}\paragraph{Formulation of GNN Models.} GNNs leverage graph topologies and node/edge features to learn a representation vector of a node, an edge or the entire graph. 
Let $\mathcal{G}=\{\mathcal{V}, \mathcal{E}\}$ denote a directed/undirected graph with nodes $v_i \in \mathcal{V}$ and edges $(v_i, v_j) \in \mathcal{E}$, where $\{v_j\}$ is the set of neighboring nodes of $v_i$.
Each node has an associated node feature $\mathcal{X} = [x_1 \ x_2 \ ... \ x_n]$.
For example, in the task of 3D object classification, $x$ can be set as the 3D coordinates.

Existing GNNs follow an iterative neighborhood aggregation scheme at each GNN layer, where each node $v_i$ iteratively gathers features from its neighboring nodes $\{v_j\}$ to capture the structural information \cite{li2019deepgcns,xu2018powerful}.
Let $\mathcal{X}_i^{\ell}$ denote the feature vector of the node $v_i$ at layer $\ell$.
The corresponding updated feature vector $\mathcal{X}_i^{\ell+1}$ in a GNN can then be formulated as:
\begin{equation}
    \vspace{-0.2em}
    \mathcal{X}_i^{\ell+1} = f \left( \mathcal{X}_i^{\ell}, \{ \mathcal{X}_j^{\ell}: (j,i)\in \mathcal{E} \} \right),
\label{eq:aggre}
\vspace{-0.2em}
\end{equation}
where $\mathcal{X}_j^{\ell}$ represents the feature associated with the neighboring nodes.
$f$ is a mapping function that takes $\mathcal{X}_i^{\ell}$ as well as $\mathcal{X}_j^{\ell}$ as inputs.
The choice of the mapping $f$ corresponds to different architectures of GNNs.

For the sake of simplicity, we take here graph convolutional network (GCN) proposed by Kipf and Welling \cite{kipf2017semi} as an example GNN architecture for the following discussions.
We denote $\text{Mean}$ as the mean aggregator that computes an average of the incoming messages and $\mathcal{W}$ as the learnable weight matrix for feature transformation.
The general GNN form in Eq.~\ref{eq:aggre} can then be instantiated for GCN as:
$
\mathcal{X}_i^{\ell+1} = \text{ReLU} \left( \mathcal{W}^l \ \text{Mean}_{(j,i)\in \mathcal{E}} \mathcal{X}_j^{\ell} \right) 
$
or 
$
\mathcal{X}_i^{\ell+1} = \text{ReLU} \left( \text{Mean}_{(j,i)\in \mathcal{E}} \mathcal{W}^l \mathcal{X}_j^{\ell} \right),
$
which respectively correspond to the case where aggregation comes first or comes after the feature transformation step~\cite{wang2019deep}.

\vspace{-1em}
\paragraph{Vanilla 1-bit GNN Models.} We develop a na\"ive binarized GNN framework to compress cumbersome GNN models, by directly binarizing 32-bit input features and learnable weights in the feature transformation step into 1-bit ones.

Specifically, for the case of vanilla binary GCN, the forward propagation process can be modeled as:
\begin{equation}
\vspace{-0.1em}
\text{Net Forward:} \quad
w_b = \text{sign}(w) =  
\left\{
\begin{aligned}
    +1 , \quad & w \geq 0  \\
    -1 , \quad & w < 0
\end{aligned}   
\right.
,
\label{eq:forward}
\vspace{-0.1em}
\end{equation}
where $w$ represents the element in the learnable weight matrix $\mathcal{W}$.
We also binarize the graph features $\mathcal{X}$ in the same manner, by replacing $w$ in Eq.~\ref{eq:forward} with the feature element $x$.

During the backward propagation, it is not feasible to simply exploit \emph{Backward Propagation (BP)} algorithm \cite{rumelhart1986learning}, as most full-precision models do, to optimize binarized graph networks, due to the undifferentiable binarization function, \ie, $\text{sign}$ in Eq.~\ref{eq:forward}.
The derivative part of the $\text{sign}$ function will lead to 0 gradients almost everywhere, thereby resulting in the vanishing gradient problem.
To alleviate this dilemma, we leverage the \emph{Straight-through Estimator (STE)} \cite{bengio2013estimating} for the backward propagation process in the binarized graph nets, formulated as:
\begin{equation}
    \text{Net Backward:} \quad
    \frac{\partial \mathcal{L}}{\partial w}
    =
    \left\{
    \begin{aligned}
        \frac{\partial \mathcal{L}}{\partial w_b} , \quad & w \in (-1, 1)  \\
        0 , \quad & \ \ \text{otherwise}
    \end{aligned}   
    \right.
    ,
    \label{eq:backward}
    \end{equation}
where $\mathcal{L}$ represents the loss function.
Essentially, Eq.~\ref{eq:backward} can be considered as propogagting the gradient through \emph{hard tanh} function, defined as: $\text{Htanh}(x)=\text{Clip}(x, -1, 1)$.

\begin{figure}[t]
    \centering
    \includegraphics[width=0.47\textwidth]{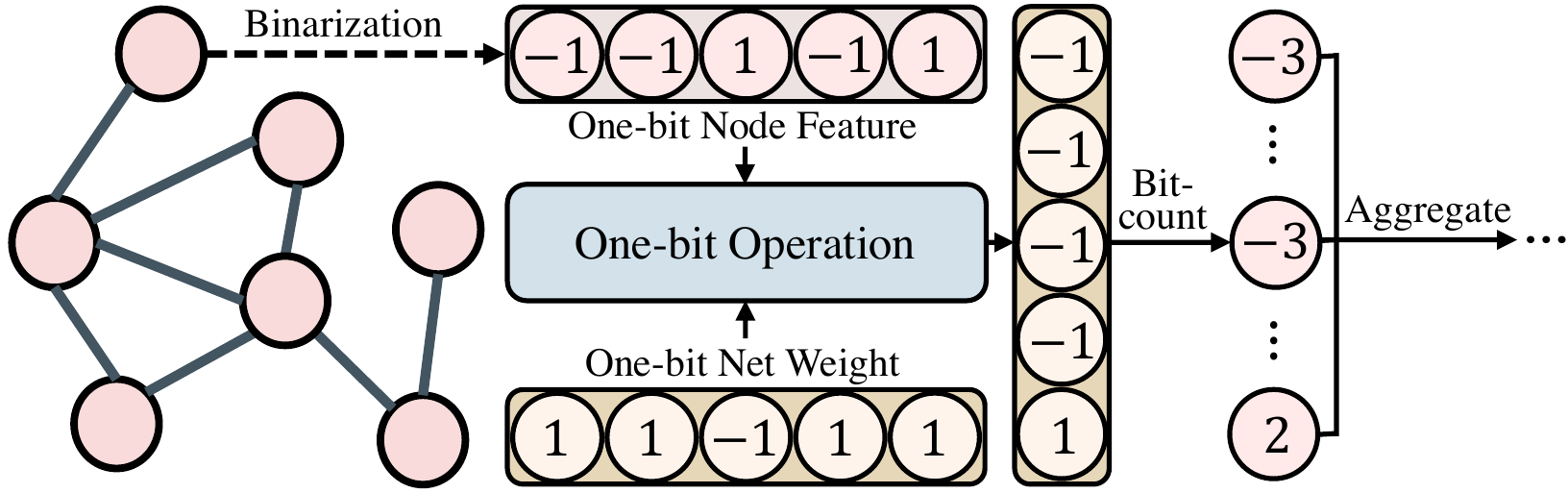}
    \caption{Illustrations of the computational workflow at an example binarized GNN layer. Despite the efficient 1-bit operations, the output features are less distinguishable between each other, leading to the challenge in the aggregation step shown in Fig.~\ref{fig:analysis}.}
    \label{fig:xnor}
\end{figure}

We illustrate in Fig.~\ref{fig:xnor} the computational workflow 
at an example binarized GCN layer for the case 
where the aggregation comes after the feature transformation.
A similar scheme can be observed for the GCN model where the aggregation happens first.
With compact node features and net weights, binarized GCN only relies on 1-bit XNOR and bit-count operations for graph-based processing, leading to an efficient and lightweight graph model that is competent for edge computing.

\vspace{-1em}\paragraph{Challenges and Possible Solutions.}

Despite the compact binarized parameters and features, we empirically observed that the results of the developed vanilla GNN were not promising as expected. 
Specifically, we conduct a preliminary experiment on the ZINC dataset \cite{irwin2012zinc} with the GCN architecture in \cite{dwivedi2020benchmarking}.
Averaged over 25 independent runs, the full-precision GCN model achieves the performance of 0.407$\pm$0.018 in terms of the mean absolute error (MAE), whereas the vanilla binarized GCN yields the result of 0.669$\pm$0.070 in MAE, which is far behind that of the full-precision one. 

We explore the reason behind this challenge of implausible performance, by looking into the internal computational process in binarized GNNs. 
Specifically, we look back on Fig.~\ref{fig:xnor}, which shows the example workflow at a binarized GCN layer where the feature transformation is performed before the aggregation step.
It is noticeable that the result of 1-bit operations lies in the discrete integer domain.
The resulted feature space is thereby much smaller than that of the 32-bit floating-point operations.  
In other words, the outputs of 1-bit operations are less distinguishable from each other.
This property, when appearing in the graph domain, leads to difficulties to extract and discriminate graph topologies in the neighborhood aggregation process, which in fact is the key to the success of graph networks.

\begin{figure}[t]
    \centering
    \includegraphics[width=0.475\textwidth]{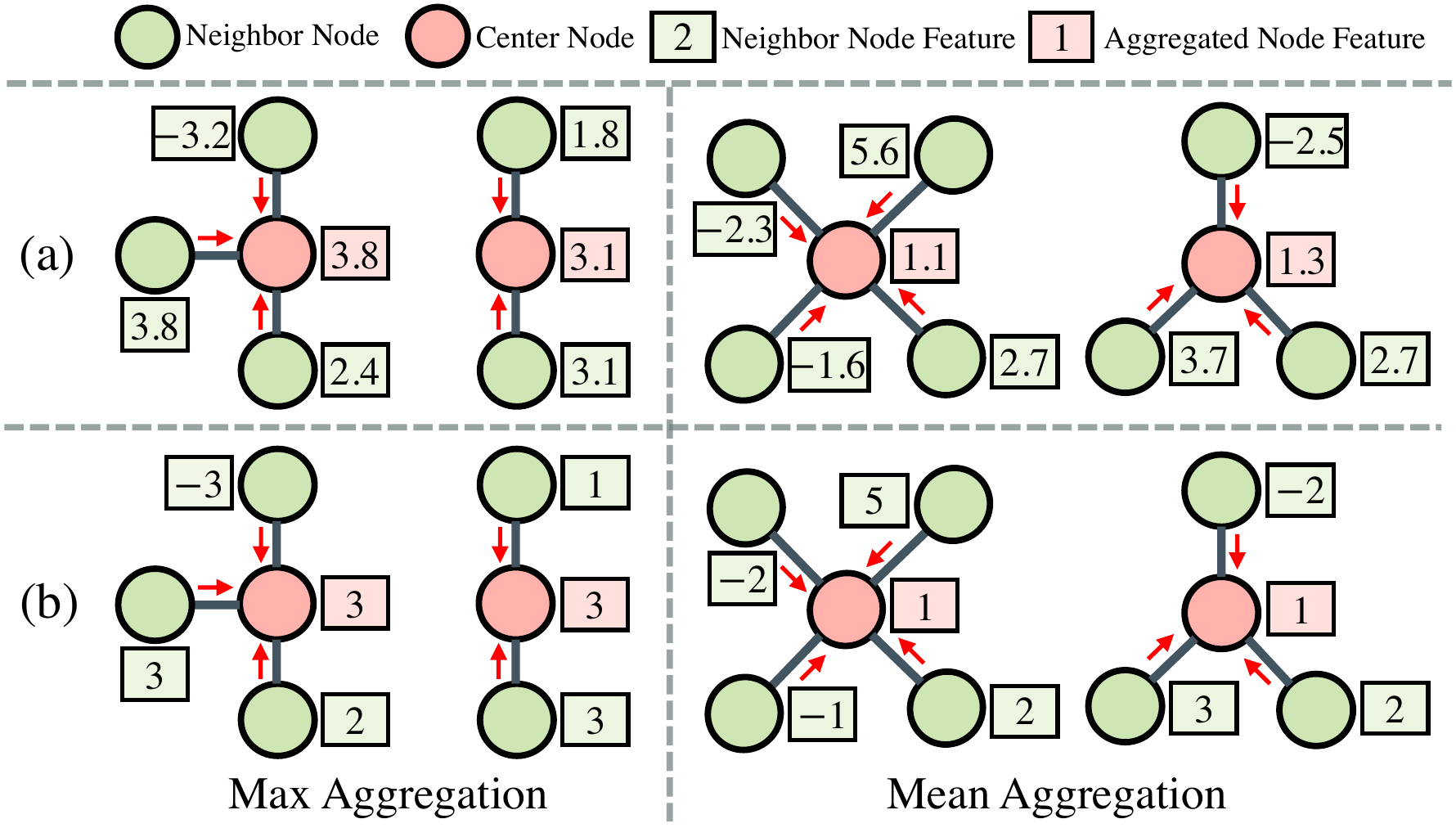}
    \caption{Example aggregation results of (a) conventional 32-bit GNN layer and (b) binarized GNN layer, corresponding to Fig.~\ref{fig:xnor}. For (a), both mean and max aggregators can distinguish the two graph structures; however, for binarized GNN (b), max and mean aggregators fail to differentiate between two topologies.}
    \label{fig:analysis}
\end{figure}

To further illustrate this dilemma, we demonstrate a couple of examples in Fig.~\ref{fig:analysis}, including both max and mean aggregation schemes that are commonly leveraged in GNNs.
Fig.~\ref{fig:analysis}(a) shows the aggregation results of the 32-bit GNN layer, where both of max and mean aggregators successfully distinguish the two different topological structures, respectively.
However, for the aggregation results of discrete integer features in binarized GNNs (Fig.~\ref{fig:analysis}(b)), neither max nor mean aggregators can discriminate the corresponding two graph structures.
Moreover, the situation will be more challenging for the case where the aggregation happens before the transformation, since the features fed into the aggregator are limited to only $1$ or $-1$.

Nevertheless, from Fig.~\ref{fig:analysis}(b), we also found that, by combining different aggregation schemes, various graph topologies could in fact become distinguishable.   
This observation motivates us to develop possible solutions to alleviate the aforementioned dilemma in vanilla binarized GNNs.
Specifically, we propose a couple of straightforward mixed multi-aggregators that combine the benefits of various aggregation schemes in two different ways.
The first one conducts multiple times of message aggregation with several different aggregators and then computes the sum over the aggregation results, leading to the performance of 0.599$\pm$0.017 in MAE with five standard aggregators.
The second one, on the other hand, concatenates the results from several independent aggregators, achieving the average result of 0.614$\pm$0.045 over 25 runs.

In spite of the improved performance, the devised possible solutions need to perform multiple times of feature aggregations at each GNN layer, resulting in heavy computational burdens. 
Motivated by this limitation, we introduce the proposed meta neighborhood aggregators, which aim to enhance the discriminative capability of topological structures and meanwhile enjoy efficient computations. 

\begin{figure*}[!t]
    \centering
    \includegraphics[width=0.85\textwidth]{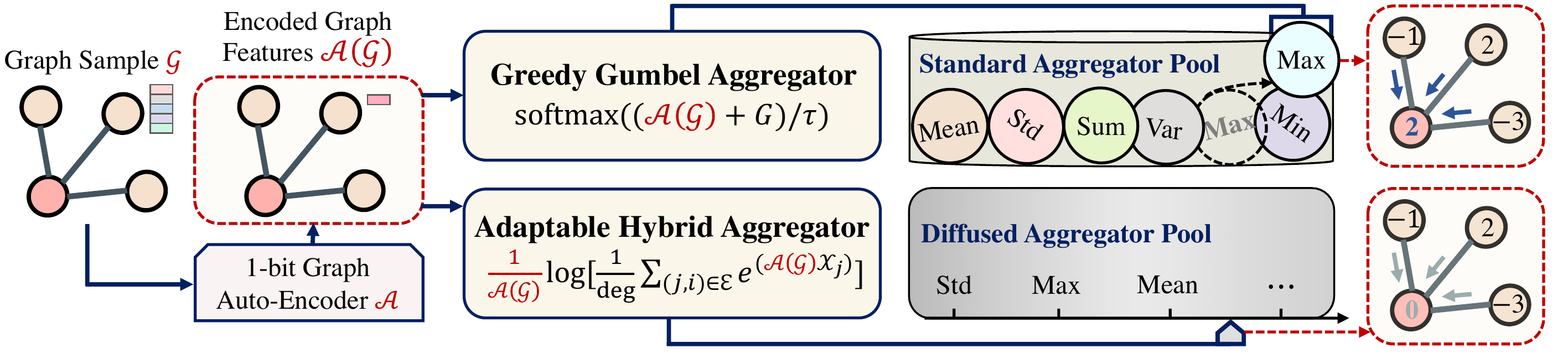}
    \caption{The overall framework of the proposed meta neighborhood aggregation methods. The upper row illustrates the workflow of the exclusive meta aggregator GNA, which 
    receives the encoded graph features from the binarized graph auto-encoder $\mathcal{A}$ (\ie, the pink trapezoid) and 
    exclusively determines a single optimal layer-wise and node-wise aggregator from a candidate aggregator pool.
    The lower row, on the other hand, demonstrates the diffused meta aggregator ANA, which amalgamates various aggregation behaviors.  
     }
    \label{fig:arch}
\end{figure*}

\section{Meta Neighborhood Aggregation}

\subsection{Overview}

Towards addressing the aforementioned limitations of the devised mixed multi-aggregators, we introduce in this section the proposed concept of the \emph{Meta Aggregator}, which aims to adaptively and efficiently adjust the way to aggregate information in a learnable manner. 
Towards this end, we propose a couple of specific forms of meta aggregators, \ie, the \emph{exclusive} meta aggregation method and the \emph{diffused} meta aggregation method, as illustrated in Fig.~\ref{fig:arch}.

The exclusive form, termed as \emph{Greedy Gumbel Neighborhood Aggregator} (GNA), learns to determine a single optimal aggregation scheme from a pool of candidate aggregators, according to the individual characteristics of the quantized graph features, as shown in the upper part of Fig.~\ref{fig:arch}.
The diffused meta form, on the other hand, adaptively learns a customized aggregation formulation that can potentially incorporate the properties of several independent aggregators, thereby termed as \emph{Adaptable Hybrid Neighborhood Aggregator} (ANA) shown in the lower part of Fig.~\ref{fig:arch}. 

In what follows, we detail the devised two forms of meta neighborhood aggregation methods, \ie, GNA and ANA, and also the associated training strategy.

{
    \setlength{\textfloatsep}{2pt}
    \renewcommand{\algorithmicrequire}{\textbf{Input:}} 
    \renewcommand{\algorithmicensure}{\textbf{Output:}}
    \begin{algorithm}[!t]
      \begin{algorithmic}[1]
        \Require{$L$: the number of layers; $\mathcal{W}$: the GNN model weight; $\mathcal{G}=\{\mathcal{V}, \mathcal{E}\}$: input graph data with nodes $v_i \in \mathcal{V}$ and edges $(v_i, v_j) \in \mathcal{E}$; $\mathcal{X}$: the input binarized node feature vector; $\mathcal{A}$: the graph auto-encoder; \emph{Meta-Aggre.}$\in$\{GNA, ANA\}: the choice of meta neighborhood aggregators.}
        \Ensure{$\mathcal{M}_{b}$: Target 1-bit binarized GNN model.}
        \For{$\ell=1$ to $L$}
            \State Feed the graph sample $\mathcal{G}$ into the GNN layer $\ell$;
            \State Binarize the GNN weight $\mathcal{W}^{\ell}$ into $\mathcal{W}_b^{\ell}$ by Eq.~\ref{eq:forward};
            \State Perform 1-bit transformation with $\mathcal{X}$ and $\mathcal{W}_b^{\ell}$;
            \State Binarize the weight $\mathcal{W}^{\mathcal{A}^{\ell}}$ of $\mathcal{A}^{\ell}$ into $\mathcal{W}^{\mathcal{A}^{\ell}}_b$ by Eq.~\ref{eq:forward};
            \State Obtain the encoded features $\mathcal{A}^{\ell}(\mathcal{G})$ with $\mathcal{W}^{\mathcal{A}^{\ell}}_b$;
            \State \emph{// Identify the choice from the two meta aggregators}
            \If{\emph{Meta-Aggre.} is GNA}
            \State \emph{// Exclusively decide an optimal aggregator}
            \State Feed $\mathcal{A}^{\ell}(\mathcal{G})$ into the GNA module.
            \State Obtain the decision $\text{GNA}_i^{\ell}$ for node $v_i$ by Eq.~\ref{equ:gna};
            \State Perform aggregations with the obtained $\text{GNA}_i^{\ell}$;
            \ElsIf{\emph{Meta-Aggre.} is ANA}
            \State \emph{// Generate a diffused aggregator}
            \State Feed $\mathcal{A}^{\ell}(\mathcal{G})$ into the ANA module;
            \State Obtain the diffused aggregator $\text{ANA}_i^{\ell}$ by Eq.~\ref{eq:ana};
            \State Perform aggregations with the obtained $\text{ANA}_i^{\ell}$;
            \EndIf
        \EndFor
        \State Optimize the binarized GNN $\mathcal{M}_{b}$ for epochs by Eq.~\ref{eq:backward}.

      \end{algorithmic}
        \caption{Training a lightweight 1-bit GNN model with the proposed meta neighborhood aggregators.}
        \label{alg:A}
    \end{algorithm}
}

%
\begin{table*}[!t]
    \caption{Results on the ZINC dataset with different architectures, in terms of the mean absolute error (MAE). From left to right: the results of the full-precision GNNs (Full), those of the 1-bit GNNs without the proposed meta aggregators (Vanilla), and the results of the 1-bit GNNs with GNA and ANA. We also provide the $p$-value of the paired $t$-test to demonstrate the statistically meaningful improvements by the proposed GNA and ANA. }
    \vspace{-0.5em}
    \footnotesize
    \begin{center}
    \setlength\tabcolsep{2 pt}
    {\renewcommand{\arraystretch}{1.2}
    \begin{tabular}{lV{2}ccccV{2}cccc}
        \noalign{\hrule height 0.8pt}
        \textbf{Methods}  & Full (GAT) \cite{velivckovic2018graph} &
        Vanilla (GAT) \cite{hubara2016binarized} & \textbf{GNA (GAT)} & \textbf{ANA (GAT)}& Full (GCN) \cite{kipf2017semi} & Vanilla (GCN) \cite{hubara2016binarized}  & \textbf{GNA (GCN)} & \textbf{ANA (GCN)} \\ 
        \hline
        \textbf{Bit-width} & 32/32 & 1/1 & \textbf{1/1} & \textbf{1/1} & 32/32 & 1/1 & \textbf{1/1} & \textbf{1/1} \\
        \textbf{Param Size}       & 399.941KB & 81.7070KB & \textbf{82.0610KB} & \textbf{81.8799KB} & 402.645KB & 82.2002KB & \textbf{82.5566KB} & \textbf{82.3740KB} \\
        \textbf{Test MAE$\pm$SD}  & 0.476$\pm$0.006 & 0.670$\pm$0.064 & \textbf{0.592$\pm$0.013} & \textbf{0.566$\pm$0.012}  & 0.407$\pm$0.018 & 0.669$\pm$0.070 & \textbf{0.608$\pm$0.024} & \textbf{0.607$\pm$0.020} \\
        \textbf{Train MAE$\pm$SD} & 0.300$\pm$0.024 & 0.610$\pm$0.066 & \textbf{0.531$\pm$0.013} & \textbf{0.453$\pm$0.019}  & 0.303$\pm$0.026 & 0.624$\pm$0.069 & \textbf{0.558$\pm$0.027} & \textbf{0.564$\pm$0.021} \\
        \hline
        $\bm{p}$\bf{-value} & \multicolumn{4}{cV{2}}{GNA vs. Vanilla: 3.010$\times$10$^{-7}$  $/$  ANA vs. Vanilla: 2.359$\times$10$^{-10}$} & \multicolumn{4}{c}{GNA vs. Vanilla: 1.597$\times$10$^{-4}$ $/$ ANA vs. Vanilla: 9.787$\times$10$^{-5}$}  \\
        \noalign{\hrule height 0.8pt}
    \end{tabular}}
\end{center}
\label{tab:zinc_1bit}
\vspace{-1.5em}
\end{table*}

\subsection{Greedy Gumbel Aggregator}

Motivated by the observation from Fig.~\ref{fig:analysis}, where different single aggregators work for a corresponding set of cases as explained in Sect.~\ref{sect:analysis}, we propose the idea of adaptively determining the optimal aggregator depending on the specific input graphs, 
as depicted in the upper part of Fig.~\ref{fig:arch}.

To this end, there are a few challenges to be addressed. 
First, the aggregation selector should understand the underlying characteristics of various input graphs without introducing much additional computational cost. 
To address this issue, we propose to leverage a 1-bit graph auto-encoder to extract meaningful information from input graphs, which is then exploited to guide the decision of different aggregation methods.


The second challenge is how to incorporate the discrete selections into the gradient descent process in training  GNNs.
One straightforward solution would be to model the discrete determination process as a state classification problem and to consider the various aggregators in the candidate pool as different labels.
However, this na\"ive attempt does not account for the uncertainty of the selector, which is likely to cause the model collapse problem where the output choice is independent of the input graphs, such as always or never picking up a specific aggregator.

To alleviate this dilemma, we propose to impose stochasticity in the aggregator decision process with greedy Gumbel sampling \cite{maddison2014sampling,veit2020convolutional} and propagate gradients through stochastic neurons through the continuous form of Gumbel-Max trick \cite{jang2016categorical}.
Specifically, we introduce such stochasticity by greedily sampling noise from the Gumbel distribution, due to its property of Gumbel-Max trick \cite{gumbel1954statistical}.
In terms of Gumbel random variables, the Gumbel-Max trick can be utilized to parameterize discrete distributions.
However, there is a argmax operation in the Gumbel-Max trick, which is not differentiable.
We thereby resort to its continuous relaxation form, termed as Gumbel-softmax estimator, to address this issue, which uses a softmax function to replace the undifferentiable argmax function.

With the aforementioned graph auto-encoder and also the Gumbel-softmax estimator to address the two challenges, respectively, the proposed greedy Gumbel aggregator (GNA) for node $v_i$ can then be formulated as: 
\begin{equation}
	\text{GNA}_i^{\ell} = \text{softmax}\Big(\big(\mathcal{A}^{\ell}(\mathcal{G}) + G\big) / \tau \Big),
    \label{equ:gna}
\end{equation}
where $\mathcal{A}^{\ell}$ represents the binarized graph auto-encoder at layer $\ell$ that extracts principal and meaningful information, and $G$ denotes the sampled Gumbel random noise. 
$\mathcal{G}$ is the input subgraph with one centered node $v_i$ and a set of its neighboring nodes $v_j$ where the connection $(v_i, v_j)\in\mathcal{E}$.
$\tau$ is a constant that denotes the temperature of the softmax.
$\text{GNA}_i^{\ell}$ is the output one-hot vector that indicates the aggregator decision at node $v_i$ and layer $\ell$ from a pool of candidate aggregators like $\{\text{max}, \text{min}, \text{std}, \text{var}, \ldots, \text{mean}\}$.

In this way, the proposed greedy Gumbel aggregator adaptively decides the optimal aggregator conditioned on each specific node and layer in a learnable manner, which can significantly improve the topological discriminative capability of the vanilla binary GNN model.

\subsection{Adaptable Hybrid Aggregator}

Despite the improved representational ability, the performance of the greedy Gumbel aggregator is bottlenecked by that of the existing standard aggregators, which leaves room for further improvement.
Motivated by this observation, we further devise an adaptable hybrid neighborhood aggregator (ANA) that can generate a hybrid form of the several standard aggregators in a learnable manner, thereby simultaneously retaining the advantages of different aggregators.
The overall computational pipeline of ANA is demonstrated in the lower part of Fig.~\ref{fig:arch}.

We start by giving the developed graph-based mathematical formulation for diffused message aggregation, defined as follows:
\begin{equation}
    \text{ANA}_i^{\ell} = \frac{1}{\mathcal{A}^{\ell}(\mathcal{G})}\log\left[\frac{1}{\text{deg}_i}\sum_{(j,i)\in \mathcal{E}}\exp(\mathcal{A}^{\ell}(\mathcal{G})\, \mathcal{X}_j^{\ell})\right]\,,
    \label{eq:ana}
\end{equation}
where $\text{deg}_i$ is the in-degree of the node $v_i$ and
$\mathcal{G}=\{\mathcal{V}, \mathcal{E}\}$ is the graph sample with edges $(v_i, v_j) \in \mathcal{E}$.
We use $\mathcal{A}^{\ell}$ to denote the 1-bit graph auto-encoder at layer $\ell$, as is also used in Eq.~\ref{equ:gna}. 
$\mathcal{X}_j^{\ell}$ represents the feature vector of the neighboring node $v_j$ at layer $\ell$, whereas
$\text{ANA}_i^{\ell}$ is the obtained diffused aggregator.

Eq.~\ref{eq:ana} can essentially approximate the max and mean functions, depending on the output of graph auto-encoder $\mathcal{A}^{\ell}(\mathcal{G})$.
Specifically, higher $\mathcal{A}^{\ell}(\mathcal{G})$ will lead to a behavior similar to that of the max aggregator, while smaller values of $\mathcal{A}^{\ell}(\mathcal{G})$ generate an effect of the mean neighborhood aggregation.
Detailed mathematical proof is provided in the supplementary material.

By slightly changing the form of Eq.~\ref{eq:ana}, we can also approximate other aggregators. 
For example, by simply adding a minus to the input graph features, Eq.~\ref{eq:ana} can approach the behavior of the min aggregation.  
Also, by utilizing the fact $\text{Var}(\mathcal{X})=\text{mean}(\mathcal{X}^2)-\big(\text{mean}(\mathcal{X})\big)^2$, the variance aggregator can be approximated by adding the square operations to Eq.~\ref{eq:ana}.
More detailed derivations and mathematical proofs can be found in the supplement.

Furthermore, it is also possible to simultaneously combine the benefits of all these approximated aggregators, by summing multiple terms in Eq.~\ref{eq:ana} with graph-based learnable weighting factors that adaptively control the diffused degree of various aggregator approximations.
We illustrate the corresponding sophisticated formulation and also more detailed explanations in the supplementary material.

\subsection{Training Strategy}

We also propose a training strategy, tailored for the proposed method.
As a whole, the principal operations of training a 1-bit GNN model with the proposed meta neighborhood aggregation approaches is concluded in Alg.~\ref{alg:A}.
For the sake of clarity, we omit the bias terms in our illustration, which have similar behavior to that of the GNN weight $\mathcal{W}$.
Also, we take the case where the feature transformation happens before the aggregation step as an example to illustrate the overall workflow. 

As can be observed from Alg.~\ref{alg:A}, at each layer, the input graph is fed into the lightweight 1-bit graph auto-encoder $\mathcal{A}$ to extract useful information that is beneficial to the following meta aggregators.
Followed by this graph encoding process, the meta neighborhood aggregation module receives the encoded features and exclusively determines an optimal aggregator, or produces a diffused aggregator that amalgamates the behaviors of several independent aggregators.
The desired 1-bit GNN model can eventually be obtained by optimizing the model for epochs with the straight-through estimator, as explained in Sect.~\ref{sect:analysis}.

\section{Experiments}
\label{sect:exp}

In this section, we perform extensive experiments on three publicly available benchmarks across diversified problem domains, including graph regression, node classification, and 3D object recognition.
Followed by the evaluations, we further provide detailed discussions regarding the strengths and weaknesses of the devised meta aggregators.

\begin{table}[t]
    \caption{Results of the proposed meta aggregation methods and other approaches for 32-bit full-precision models on the ZINC dataset, in terms of MAE. The results are averaged over 25 independent runs with 25 different random seeds.}
    \vspace{-0.5em}
    \footnotesize
    \begin{center}
    \setlength\tabcolsep{4 pt}
    {\renewcommand{\arraystretch}{1.2}
    \begin{tabular}{lV{2}ccc}
        \noalign{\hrule height 0.8pt}
        \textbf{Methods} & \textbf{Param Size} & \textbf{Test MAE$\pm$SD} & \textbf{Train MAE$\pm$SD} \\
        \noalign{\hrule height 0.5pt}
        GatedGCN \cite{bresson2017residual} & 413.027KB & 0.426$\pm$0.012 & 0.272$\pm$0.023 \\ 
        GraphSage \cite{hamilton2017inductive} & 371.004KB & 0.475$\pm$0.007 & 0.296$\pm$0.030\\
        GIN \cite{xu2018powerful} & 402.652KB & 0.387$\pm$0.019 & 0.319$\pm$0.020\\
        MoNet \cite{monti2017geometric} & 414.070KB & 0.386$\pm$0.009 & 0.299$\pm$0.016\\
        GCN \cite{kipf2017semi} & 402.645KB & 0.407$\pm$0.018 & 0.303$\pm$0.026\\
        GAT \cite{velivckovic2018graph} & 399.941KB & 0.476$\pm$0.006 & 0.300$\pm$0.024\\
        \textbf{GNA (Ours)} & \textbf{411.270KB} & \textbf{0.337$\pm$0.021} & \textbf{0.160$\pm$0.026}\\
        \textbf{ANA (Ours)} & \textbf{404.504KB} & \textbf{0.325$\pm$0.015} & \textbf{0.109$\pm$0.014}\\
        \noalign{\hrule height 0.8pt}
    \end{tabular}}
\end{center}
\label{tab:zinc_32bit}
\vspace{-1.5em}
\end{table}

\subsection{Experimental Settings}

\noindent\textbf{Datasets.} We validate the effectiveness of the proposed meta aggregation methods on three different datasets, each of which specializes in a distinct task.
Specifically, for the task of graph regression, we use the ZINC dataset \cite{irwin2012zinc}, which is one of the most popular real-world molecular datasets \cite{dwivedi2020benchmarking}.
The goal of ZINC is to regress a specific molecular property, \ie the constrained solubility, which is a critical property for developing GNNs for molecules \cite{you2018graph}.

Also, for the node classification task, we adopt the protein-protein interaction (PPI) dataset \cite{zitnik2017predicting}, which is a multi-label dataset with 24 graphs corresponding to different human tissues.
Each node in the PPI dataset is labeled with various protein functions.
The objective of PPI is thereby to predict the 121 protein functions from the interactions of human tissue proteins.
Furthermore, we utilize ModelNet40 \cite{wu20153d} for the evaluation on the task of 3D object classification.
ModelNet40 is a popular dataset for 3D object analysis \cite{qi2017pointnet,qi2017pointnet++}, containing 12,311 meshed CAD models from 40 shape categories in total.
Each object comprises a set of 3D points, with the 3D coordinates as the features. 
The goal is to predict the category of each 3D shape.

\vspace{-1em}\paragraph{Implementation Details.} 
We primarily use three heterogeneous architectures, including graph convolutional network (GCN) \cite{kipf2017semi}, graph attention network (GAT) \cite{velivckovic2018graph}, as well as dynamic graph convolutional model (DGCNN) \cite{dgcnn} to evaluate the proposed meta aggregation approach.
For other settings such as learning rates and batch size, we follow those in the works of \cite{dwivedi2020benchmarking}, \cite{velivckovic2018graph}, and \cite{dgcnn} for the tasks of graph regression, node classification, and point cloud classification, respectively.

In particular, for more convincing evaluations, we report the results on the ZINC dataset over 25 independent runs with 25 different random seeds.
Also, as done in the field of CNN binarization \cite{qin2020binary}, we keep the first and the last GNN layer full-precision and binarize the other GNN layers for all the comparison methods.
More detailed task-by-task architecture designs as well as the hyperparameter settings can be found in the supplementary material.

\begin{table}[t]
    \caption{Results on the PPI dataset for the task of node classification, in terms of micro-averaged F$_1$ score. Detailed network architectures can be found in the supplementary material.}
    \vspace{-0.5em}
    \footnotesize
    \begin{center}
    \setlength\tabcolsep{6 pt}
    {\renewcommand{\arraystretch}{1.2}
    \begin{tabular}{lV{2}ccc}
        \noalign{\hrule height 0.8pt}
        \textbf{Methods} & \textbf{Bit-width} & \textbf{Param Size} & \textbf{$\text{F}_1$ Score} \\
        \noalign{\hrule height 0.5pt}
        Full Prec. \cite{velivckovic2018graph} & 32/32 & 43.7712MB & 98.70 \\
        \hline
        Vanilla \cite{hubara2016binarized} & 1/1 & 28.2560MB & 92.68\\
        \textbf{GNA (Ours)} & \textbf{1/1} & \textbf{28.2572MB} & \textbf{97.52}\\
        \textbf{ANA (Ours)} & \textbf{1/1} & \textbf{28.2565MB} & \textbf{97.71} \\
        \noalign{\hrule height 0.8pt}
    \end{tabular}}
\end{center}
\label{tab:ppi}
\vspace{-2em}
\end{table}
%

\begin{figure*}[!t]
    \centering
    \includegraphics[width=0.7\textwidth]{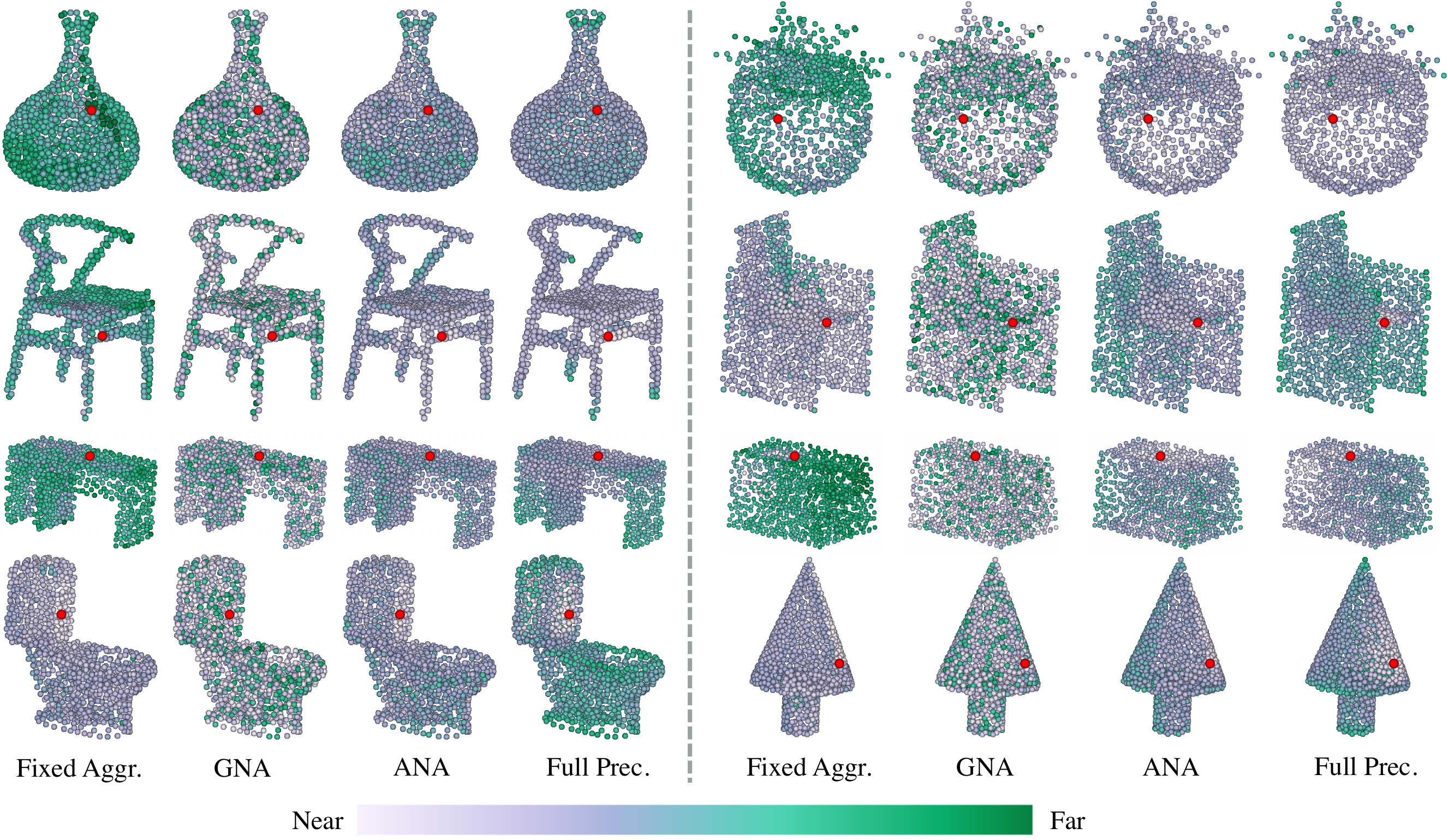}\\
    \caption{Visualization results of the learned feature space, depicted as the distance between the red point and the rest of the others. The visualized features are extracted from the intermediate layer of the models. More results can be found in the supplementary material.}
    \label{fig:quality}
    \vspace{-0.5em}
    \end{figure*}

\subsection{Results}

\noindent\textbf{Graph Regression.} Tab.~\ref{tab:zinc_1bit} shows the ablation results of the vanilla 1-bit GNN models and those of GNNs with the proposed meta neighborhood aggregators GAN and ANA.
Specifically, we report the results on two GNN architectures, \ie, GCN \cite{kipf2017semi} and GAT \cite{velivckovic2018graph}, by averaging over 25 independent runs with 25 seeds.

The proposed GNA and ANA, as shown in Tab.~\ref{tab:zinc_1bit}, achieves gratifying performance in terms of both test and train MAE, and at the same time maintains a compact model size.
Moreover, we provide in the last row of Tab.~\ref{tab:zinc_1bit} the $p$-value of the paired $t$-test between the 1-bit GNNs with a fixed aggregator (Vanilla) and those with the proposed learnable meta aggregators.
The corresponding results statistically validate the effectiveness of the proposed method.

Furthermore, we show in Tab.~\ref{tab:zinc_32bit} the results of extending the proposed meta aggregators to full-precision GNNs and compare them with those of the state-of-the-art approaches \cite{bresson2017residual,hamilton2017inductive,xu2018powerful,monti2017geometric,kipf2017semi,velivckovic2018graph}.
Specifically, the results in the last two rows of Tab.~\ref{tab:zinc_32bit} are obtained by simply replacing the pre-defined aggregator in GAT with the proposed GNA and ANA.
As can be observed from Tab.~\ref{tab:zinc_32bit}, the proposed method outperforms other approaches by a large margin, and meanwhile introduces few additional parameters. 

\noindent\textbf{Node Classification.} In Tab.~\ref{tab:ppi}, we demonstrate the results of different methods with the GAT architecture.
The proposed GNA and ANA, as shown in Tab.~\ref{tab:ppi}, yield results on par with those of the 32-bit full-precision models, but comes with a more lightweight architecture.
The proposed method also outperforms the vanilla 1-bit GNN model that relies on a fixed aggregation scheme.

\begin{table}[t]
    \caption{Results on the ModelNet40 dataset for 3D object recognition, in terms of the overall accuracy (Acc) and the mean class accuracy (mAcc).}
    \vspace{-0.5em}
    \footnotesize
    \begin{center}
    \setlength\tabcolsep{5 pt}
    {\renewcommand{\arraystretch}{1.2}
    \begin{tabular}{lV{2}cccc}
        \noalign{\hrule height 0.8pt}
        \textbf{Methods} & \textbf{Bit-width} & \textbf{Param Size} & \textbf{Acc (\%)} & \textbf{mAcc (\%)} \\
        \noalign{\hrule height 0.5pt}
        Full Prec. \cite{dgcnn} & 32/32 & 1681.66KB & 92.42 & 89.51 \\
        \hline
        Vanilla \cite{hubara2016binarized} & 1/1 & 1091.20KB & 74.19 & 65.95  \\
        \textbf{GNA (Ours)} & \textbf{1/1} & \textbf{1091.30KB} & \textbf{78.36} & \textbf{71.67} \\
        \textbf{ANA (Ours)} & \textbf{1/1} & \textbf{1091.30KB} & \textbf{84.64} & \textbf{78.89} \\
        \noalign{\hrule height 0.8pt}
    \end{tabular}}
\end{center}
\label{tab:dgcnn}
\vspace{-2.5em}
\end{table}

\vspace{-1.5em}\paragraph{3D Object Recognition.} 
The results of the proposed approach and other methods on the ModelNet40 dataset are shown in Tab.~\ref{tab:dgcnn}.
We build our network here based on the architecture designed in \cite{yang2020distillating}.
We also demonstrate in Fig.~\ref{fig:quality} the corresponding visualization results of different approaches, where the column termed as ``Fixed Aggr.'' in Fig.~\ref{fig:quality} corresponds to the ``Vanilla'' model in Tab.~\ref{tab:dgcnn}. 
With the proposed meta aggregation schemes, the 1-bit GNN model gains a boost by more than 10\% in both the overall accuracy and the mean class accuracy.
This improvement is also illustrated in Fig.~\ref{fig:quality}, where the proposed meta aggregators help the 1-bit GNN learn a closer structure to that of the full-precision GNN model.

\subsection{Discussions}

We provide here a detailed account of the strengths and weaknesses of the proposed two meta aggregators GNA and ANA.
For the exclusive meta form GNA, the performance can potentially be further enhanced with the advance of novel aggregation schemes.
In other words, the results of GNA depend on those of every single aggregator in the candidate aggregation pool, which at the same time is a weakness of GNA since its performance is bottlenecked by that of the single aggregator.
The diffused form ANA, on the other hand, may simultaneously retain the benefits of several popular aggregators.
However, the mathematical form in Eq.~\ref{eq:ana} limits the type of aggregators that ANA can potentially approximate,
meaning that ANA may not have much room for further improvement even with the emergence of novel and prevailing aggregators in the future.

\section{Conclusions}

In this paper, we propose a couple of learnable aggregation schemes for 1-bit compact GNNs.
The goal of the proposed method is to enhance the topological discriminative ability of the 1-bit GNNs.
This is achieved by adaptively selecting a single aggregator, or generating a hybrid aggregation form that can simultaneously maintain the strengths of several aggregators.
Moreover, the proposed meta aggregation schemes can be readily extended to 
the full-precision GNN models.
Experiments across various domains demonstrate that, with the proposed meta aggregators, the 1-bit GNN   yields results on par with those of the cumbersome full-precision ones.
In our future work, we will strive to generalize the proposed aggregator to compact and lightweight visual transformers.

\vspace{-1.5em}
\paragraph{Acknowledgements.} 
Mr Yongcheng Jing is supported by ARC FL-170100117.
Xinchao Wang is supported by the Start-up Fund of National University of Singapore.

{\small
\bibliographystyle{ieee_fullname}
\bibliography{MYRE}

\begin{thebibliography}{10}\itemsep=-1pt

\bibitem{bahri2020binary}
Mehdi Bahri, Ga{\'e}tan Bahl, and Stefanos Zafeiriou.
\newblock Binary graph neural networks.
\newblock {\em arXiv preprint arXiv:2012.15823}, 2020.

\bibitem{bengio2013estimating}
Yoshua Bengio, Nicholas L{\'e}onard, and Aaron Courville.
\newblock Estimating or propagating gradients through stochastic neurons for
  conditional computation.
\newblock {\em arXiv preprint arXiv:1308.3432}, 2013.

\bibitem{bresson2017residual}
Xavier Bresson and Thomas Laurent.
\newblock Residual gated graph convnets.
\newblock {\em arXiv preprint arXiv:1711.07553}, 2017.

\bibitem{bulat2019xnor}
Adrian Bulat and Georgios Tzimiropoulos.
\newblock Xnor-net++: Improved binary neural networks.
\newblock In {\em BMVC}, 2019.

\bibitem{chen2020addernet}
Hanting Chen, Yunhe Wang, Chunjing Xu, Boxin Shi, Chao Xu, Qi Tian, and Chang
  Xu.
\newblock Addernet: Do we really need multiplications in deep learning?
\newblock In {\em CVPR}, 2020.

\bibitem{courbariaux2015binaryconnect}
Matthieu Courbariaux, Yoshua Bengio, and Jean-Pierre David.
\newblock Binaryconnect: training deep neural networks with binary weights
  during propagations.
\newblock In {\em NeurIPS}, 2015.

\bibitem{dwivedi2020benchmarking}
Vijay~Prakash Dwivedi, Chaitanya~K Joshi, Thomas Laurent, Yoshua Bengio, and
  Xavier Bresson.
\newblock Benchmarking graph neural networks.
\newblock {\em arXiv preprint arXiv:2003.00982}, 2020.

\bibitem{gu2018learning}
Jiayuan Gu, Han Hu, Liwei Wang, Yichen Wei, and Jifeng Dai.
\newblock Learning region features for object detection.
\newblock In {\em ECCV}, 2018.

\bibitem{gumbel1954statistical}
Emil~Julius Gumbel.
\newblock {\em Statistical theory of extreme values and some practical
  applications: a series of lectures}.
\newblock US Government Printing Office, 1954.

\bibitem{hamilton2017inductive}
Will Hamilton, Zhitao Ying, and Jure Leskovec.
\newblock Inductive representation learning on large graphs.
\newblock In {\em NeurIPS}, 2017.

\bibitem{hu2018relation}
Han Hu, Jiayuan Gu, Zheng Zhang, Jifeng Dai, and Yichen Wei.
\newblock Relation networks for object detection.
\newblock In {\em CVPR}, 2018.

\bibitem{hu2020open}
Weihua Hu, Matthias Fey, Marinka Zitnik, Yuxiao Dong, Hongyu Ren, Bowen Liu,
  Michele Catasta, and Jure Leskovec.
\newblock Open graph benchmark: Datasets for machine learning on graphs.
\newblock {\em arXiv preprint arXiv:2005.00687}, 2020.

\bibitem{huang2018adaptive}
Wenbing Huang, Tong Zhang, Yu Rong, and Junzhou Huang.
\newblock Adaptive sampling towards fast graph representation learning.
\newblock In {\em NeurIPS}, 2018.

\bibitem{hubara2016binarized}
Itay Hubara, Matthieu Courbariaux, Daniel Soudry, Ran El-Yaniv, and Yoshua
  Bengio.
\newblock Binarized neural networks.
\newblock In {\em NeurIPS}, 2016.

\bibitem{hubara2016binarizedarxiv}
Itay Hubara, Matthieu Courbariaux, Daniel Soudry, Ran El-Yaniv, and Yoshua
  Bengio.
\newblock Binarized neural networks: Training neural networks with weights and
  activations constrained to+ 1 or-1.
\newblock {\em arXiv preprint arXiv:1602.02830}, 2016.

\bibitem{irwin2012zinc}
John~J Irwin, Teague Sterling, Michael~M Mysinger, Erin~S Bolstad, and Ryan~G
  Coleman.
\newblock Zinc: a free tool to discover chemistry for biology.
\newblock {\em Journal of chemical information and modeling}, 2012.

\bibitem{jain2016structural}
Ashesh Jain, Amir~R Zamir, Silvio Savarese, and Ashutosh Saxena.
\newblock Structural-rnn: Deep learning on spatio-temporal graphs.
\newblock In {\em CVPR}, 2016.

\bibitem{jang2016categorical}
Eric Jang, Shixiang Gu, and Ben Poole.
\newblock Categorical reparameterization with gumbel-softmax.
\newblock {\em arXiv preprint arXiv:1611.01144}, 2016.

\bibitem{jing2021amalgamating}
Yongcheng Jing, Yiding Yang, Xinchao Wang, Mingli Song, and Dacheng Tao.
\newblock Amalgamating knowledge from heterogeneous graph neural networks.
\newblock In {\em CVPR}, 2021.

\bibitem{kim2016bitwise}
Minje Kim and Paris Smaragdis.
\newblock Bitwise neural networks.
\newblock {\em arXiv preprint arXiv:1601.06071}, 2016.

\bibitem{kipf2017semi}
Thomas~N Kipf and Max Welling.
\newblock Semi-supervised classification with graph convolutional networks.
\newblock In {\em ICLR}, 2017.

\bibitem{landrieu2018large}
Loic Landrieu and Martin Simonovsky.
\newblock Large-scale point cloud semantic segmentation with superpoint graphs.
\newblock In {\em CVPR}, 2018.

\bibitem{li2019deepgcns}
Guohao Li, Matthias Muller, Ali Thabet, and Bernard Ghanem.
\newblock Deepgcns: Can gcns go as deep as cnns?
\newblock In {\em ICCV}, 2019.

\bibitem{li2020deepergcn}
Guohao Li, Chenxin Xiong, Ali Thabet, and Bernard Ghanem.
\newblock Deepergcn: All you need to train deeper gcns.
\newblock {\em arXiv preprint arXiv:2006.07739}, 2020.

\bibitem{li2018deeper}
Qimai Li, Zhichao Han, and Xiao-Ming Wu.
\newblock Deeper insights into graph convolutional networks for semi-supervised
  learning.
\newblock In {\em AAAI}, 2018.

\bibitem{li2018adaptive}
Ruoyu Li, Sheng Wang, Feiyun Zhu, and Junzhou Huang.
\newblock Adaptive graph convolutional neural networks.
\newblock In {\em AAAI}, 2018.

\bibitem{liu2021overcoming}
Huihui Liu, Yiding Yang, and Xinchao Wang.
\newblock Overcoming catastrophic forgetting in graph neural networks.
\newblock In {\em AAAI}, 2021.

\bibitem{ma2019disentangled}
Jianxin Ma, Peng Cui, Kun Kuang, Xin Wang, and Wenwu Zhu.
\newblock Disentangled graph convolutional networks.
\newblock In {\em ICML}, 2019.

\bibitem{maddison2014sampling}
Chris~J Maddison, Daniel Tarlow, and Tom Minka.
\newblock A* sampling.
\newblock In {\em NeurIPS}, 2014.

\bibitem{martinez2020training}
Brais Martinez, Jing Yang, Adrian Bulat, and Georgios Tzimiropoulos.
\newblock Training binary neural networks with real-to-binary convolutions.
\newblock In {\em ICLR}, 2020.

\bibitem{milz2018visual}
Stefan Milz, Georg Arbeiter, Christian Witt, Bassam Abdallah, and Senthil
  Yogamani.
\newblock Visual slam for automated driving: Exploring the applications of deep
  learning.
\newblock In {\em CVPR Workshop}, 2018.

\bibitem{monti2017geometric}
Federico Monti, Davide Boscaini, Jonathan Masci, Emanuele Rodola, Jan Svoboda,
  and Michael~M Bronstein.
\newblock Geometric deep learning on graphs and manifolds using mixture model
  cnns.
\newblock In {\em CVPR}, 2017.

\bibitem{nt2019revisiting}
Hoang NT and Takanori Maehara.
\newblock Revisiting graph neural networks: All we have is low-pass filters.
\newblock {\em arXiv preprint arXiv:1905.09550}, 2019.

\bibitem{pellegrini2021learning}
Giovanni Pellegrini, Alessandro Tibo, Paolo Frasconi, Andrea Passerini, and
  Manfred Jaeger.
\newblock Learning aggregation functions.
\newblock In {\em IJCAI}, 2021.

\bibitem{qi2017pointnet}
Charles~R Qi, Hao Su, Kaichun Mo, and Leonidas~J Guibas.
\newblock Pointnet: Deep learning on point sets for 3d classification and
  segmentation.
\newblock In {\em CVPR}, 2017.

\bibitem{qi2017pointnet++}
Charles~Ruizhongtai Qi, Li Yi, Hao Su, and Leonidas~J Guibas.
\newblock Pointnet++: Deep hierarchical feature learning on point sets in a
  metric space.
\newblock In {\em NeurIPS}, 2017.

\bibitem{qi2018learning}
Siyuan Qi, Wenguan Wang, Baoxiong Jia, Jianbing Shen, and Song-Chun Zhu.
\newblock Learning human-object interactions by graph parsing neural networks.
\newblock In {\em ECCV}, 2018.

\bibitem{qi20173d}
Xiaojuan Qi, Renjie Liao, Jiaya Jia, Sanja Fidler, and Raquel Urtasun.
\newblock 3d graph neural networks for rgbd semantic segmentation.
\newblock In {\em ICCV}, 2017.

\bibitem{qin2020binary}
Haotong Qin, Ruihao Gong, Xianglong Liu, Xiao Bai, Jingkuan Song, and Nicu
  Sebe.
\newblock Binary neural networks: A survey.
\newblock {\em Pattern Recognition}, 2020.

\bibitem{rastegari2016xnor}
Mohammad Rastegari, Vicente Ordonez, Joseph Redmon, and Ali Farhadi.
\newblock Xnor-net: Imagenet classification using binary convolutional neural
  networks.
\newblock In {\em ECCV}, 2016.

\bibitem{rumelhart1986learning}
David~E Rumelhart, Geoffrey~E Hinton, and Ronald~J Williams.
\newblock Learning representations by back-propagating errors.
\newblock {\em Nature}, 1986.

\bibitem{sarlin2020superglue}
Paul-Edouard Sarlin, Daniel DeTone, Tomasz Malisiewicz, and Andrew Rabinovich.
\newblock Superglue: Learning feature matching with graph neural networks.
\newblock In {\em CVPR}, 2020.

\bibitem{scarselli2008graph}
Franco Scarselli, Marco Gori, Ah~Chung Tsoi, Markus Hagenbuchner, and Gabriele
  Monfardini.
\newblock The graph neural network model.
\newblock {\em TNN}, 2008.

\bibitem{shen2019amalgamating}
Chengchao Shen, Xinchao Wang, Jie Song, Li Sun, and Mingli Song.
\newblock Amalgamating knowledge towards comprehensive classification.
\newblock In {\em AAAI}, 2019.

\bibitem{shen2021progressive}
Chengchao Shen, Xinchao Wang, Youtan Yin, Jie Song, Sihui Luo, and Mingli Song.
\newblock Progressive network grafting for few-shot knowledge distillation.
\newblock In {\em AAAI}, 2021.

\bibitem{shen2019customizing}
Chengchao Shen, Mengqi Xue, Xinchao Wang, Jie Song, Li Sun, and Mingli Song.
\newblock Customizing student networks from heterogeneous teachers via adaptive
  knowledge amalgamation.
\newblock In {\em ICCV}, 2019.

\bibitem{veit2020convolutional}
Andreas Veit and Serge Belongie.
\newblock Convolutional networks with adaptive inference graphs.
\newblock {\em IJCV}, 2020.

\bibitem{velivckovic2018graph}
Petar Veli{\v{c}}kovi{\'c}, Guillem Cucurull, Arantxa Casanova, Adriana Romero,
  Pietro Lio, and Yoshua Bengio.
\newblock Graph attention networks.
\newblock In {\em ICLR}, 2018.

\bibitem{wang2020binarized}
Hanchen Wang, Defu Lian, Ying Zhang, Lu Qin, Xiangjian He, Yiguang Lin, and
  Xuemin Lin.
\newblock Binarized graph neural network.
\newblock {\em arXiv preprint arXiv:2004.11147}, 2020.

\bibitem{wang2018graphgan}
Hongwei Wang, Jia Wang, Jialin Wang, Miao Zhao, Weinan Zhang, Fuzheng Zhang,
  Xing Xie, and Minyi Guo.
\newblock Graphgan: Graph representation learning with generative adversarial
  nets.
\newblock In {\em AAAI}, 2018.

\bibitem{wang2020bi}
Junfu Wang, Yunhong Wang, Zhen Yang, Liang Yang, and Yuanfang Guo.
\newblock Bi-gcn: Binary graph convolutional network.
\newblock {\em arXiv preprint arXiv:2010.07565}, 2020.

\bibitem{wang2019deep}
Minjie Wang, Lingfan Yu, Da Zheng, Quan Gan, Yu Gai, Zihao Ye, Mufei Li,
  Jinjing Zhou, Qi Huang, Chao Ma, et~al.
\newblock Deep graph library: Towards efficient and scalable deep learning on
  graphs.
\newblock In {\em ICLR Workshop}, 2019.

\bibitem{wang2019dynamic}
Yue Wang, Yongbin Sun, Ziwei Liu, Sanjay~E Sarma, Michael~M Bronstein, and
  Justin~M Solomon.
\newblock Dynamic graph cnn for learning on point clouds.
\newblock {\em TOG}, 2019.

\bibitem{dgcnn}
Yue Wang, Yongbin Sun, Ziwei Liu, Sanjay~E. Sarma, Michael~M. Bronstein, and
  Justin~M. Solomon.
\newblock Dynamic graph cnn for learning on point clouds.
\newblock {\em TOG}, 2019.

\bibitem{wu2020comprehensive}
Zonghan Wu, Shirui Pan, Fengwen Chen, Guodong Long, Chengqi Zhang, and S~Yu
  Philip.
\newblock A comprehensive survey on graph neural networks.
\newblock {\em TNNLS}, 2020.

\bibitem{wu20153d}
Zhirong Wu, Shuran Song, Aditya Khosla, Fisher Yu, Linguang Zhang, Xiaoou Tang,
  and Jianxiong Xiao.
\newblock 3d shapenets: A deep representation for volumetric shapes.
\newblock In {\em CVPR}, 2015.

\bibitem{xu2018powerful}
Keyulu Xu, Weihua Hu, Jure Leskovec, and Stefanie Jegelka.
\newblock How powerful are graph neural networks?
\newblock In {\em ICLR}, 2019.

\bibitem{yan2018spatial}
Sijie Yan, Yuanjun Xiong, and Dahua Lin.
\newblock Spatial temporal graph convolutional networks for skeleton-based
  action recognition.
\newblock In {\em AAAI}, 2018.

\bibitem{yang2020factorizable}
Yiding Yang, Zunlei Feng, Mingli Song, and Xinchao Wang.
\newblock Factorizable graph convolutional networks.
\newblock In {\em NeurIPS}, 2020.

\bibitem{yang2020distillating}
Yiding Yang, Jiayan Qiu, Mingli Song, Dacheng Tao, and Xinchao Wang.
\newblock Distilling knowledge from graph convolutional networks.
\newblock In {\em CVPR}, 2020.

\bibitem{yang2021learning}
Yiding Yang, Zhou Ren, Haoxiang Li, Chunluan Zhou, Xinchao Wang, and Gang Hua.
\newblock Learning dynamics via graph neural networks for human pose estimation
  and tracking.
\newblock In {\em CVPR}, 2021.

\bibitem{yang2021spagan}
Yiding Yang, Xinchao Wang, Mingli Song, Junsong Yuan, and Dacheng Tao.
\newblock Spagan: Shortest path graph attention network.
\newblock In {\em IJCAI}, 2019.

\bibitem{you2018graph}
Jiaxuan You, Bowen Liu, Rex Ying, Vijay Pande, and Jure Leskovec.
\newblock Graph convolutional policy network for goal-directed molecular graph
  generation.
\newblock In {\em NeurIPS}, 2018.

\bibitem{yu2017compressing}
Xiyu Yu, Tongliang Liu, Xinchao Wang, and Dacheng Tao.
\newblock On compressing deep models by low rank and sparse decomposition.
\newblock In {\em CVPR}, 2017.

\bibitem{zhou2018graph}
Jie Zhou, Ganqu Cui, Zhengyan Zhang, Cheng Yang, Zhiyuan Liu, Lifeng Wang,
  Changcheng Li, and Maosong Sun.
\newblock Graph neural networks: A review of methods and applications.
\newblock {\em arXiv preprint arXiv:1812.08434}, 2018.

\bibitem{zitnik2017predicting}
Marinka Zitnik and Jure Leskovec.
\newblock Predicting multicellular function through multi-layer tissue
  networks.
\newblock {\em Bioinformatics}, 2017.

\end{thebibliography}
}

\end{document}